\documentclass[11pt,a4paper]{article}
\usepackage[hyperref]{emnlp2018}
\usepackage{times}
\usepackage{latexsym}
\usepackage[skip=6pt]{caption}
\usepackage{url}

\aclfinalcopy 

\usepackage{graphicx}
\usepackage{multirow}

\title{Argument Component Classification for Classroom Discussions}

\author{Luca Lugini \and Diane Litman\\
  Computer Science Department \& Learning Research and Development Center\\
  University of Pittsburgh \\
  Pittsburgh, PA 15260 \\
  {\tt \{lul32,dlitman\}@pitt.edu}}

\date{}

\begin{document}
\maketitle
\begin{abstract}
This paper focuses on argument component classification for transcribed spoken classroom discussions, with the goal of automatically classifying student utterances into claims, evidence, and warrants.
We show that an existing method for argument component classification developed for another educationally-oriented domain performs poorly on our dataset.
We then show that feature sets from prior work on argument mining for student essays and online dialogues can be used to improve performance considerably.
We also provide a comparison between convolutional neural networks and recurrent neural networks when trained under different conditions to classify argument components in classroom discussions.
While neural network models are not always able to outperform a logistic regression model, we were able to gain some useful insights: convolutional networks are more robust than recurrent networks both at the character and at the word level, and specificity information can help boost performance in multi-task training.
\end{abstract}

\section{Introduction}
\label{sec:introduction}
Although there is no universally agreed upon definition, argument mining is an area of natural language processing which aims to extract structured knowledge from free-form unstructured language.
In particular, argument mining systems are built with goals such as: detecting what parts of a text express an argument component, known as argument component identification; categorizing arguments into different component types (e.g. claim, evidence), known as argument component classification;
understanding if/how different components are connected to form an argumentative structure (e.g. using evidence to support/attack a claim), known as argument relation identification.
The development and release to the public of corpora and annotations in recent years have 
contributed to the increasing interest in the area.

One domain in which argument mining is rarely found in the literature is educational discussions.
Classroom discussions are a part of students' daily life, and they are a common pedagogical approach for enhancing student skills.
For example, student-centered classroom discussions are an important contributor to the development of students' reading, writing, and reasoning skills in the context of English Language Arts (ELA) classes \cite{Applebee:03, Reznitskaya:13}. This impact is  reflected in students' problem solving and disciplinary skills \cite{Engle:02,Murphy:09,Elizabeth:12}.
With the increasing importance of argumentation in classrooms, especially in the context of student-centered discussions, automatically performing argument component classification is a  first step for building tools aimed at helping teachers analyze and better understand student arguments, with the goal of improving students' learning outcomes.

Many current argument mining systems focus on analyzing argumentation  in student essays \cite{Stab:14,Stab:17,Nguyen:15,Nguyen:18}, online dialogues \cite{Swanson:15,Mclaren:10,Ghosh:14,Lawrence:17}, or in the legal domain \cite{Ashley:13,Palau:09}.
A key difference between these studies and our work consists in the source of linguistic content: although we analyze written transcriptions of discussions, the original source for our corpora consists of spoken, multi-party, educational discussions, and the difference in cognitive skills and grammatical structure between written and spoken language \cite{Biber:88,Chafe:87} introduces additional complexity.


Our work and previous research studies on student essays share the trait of analyzing argumentation in an educational context.
However, while student essays are typically written by an individual student, in classroom discussions arguments are formed collaboratively between multiple parties (i.e. multiple students and possibly teachers).
While our work shares the multi-party context in which arguments are made with research aimed at argument mining in online dialogues, prior online dialogue studies have not been contextualized in the educational domain.

Given these differences, we believe that argument mining models  for student essays and online dialogues will perform poorly when directly applied to educational discussions.
However, since similarities between the domains do exist, we expect that features exploited by such argument mining models can help us in classifying argument components in classroom discussions.
Moreover, unlike the other two domains, we have access to labels belonging to a different (but related) class, specificity, which we can try to incorporate in argumentation models to boost performance.


Our contributions are as follows. We first experimentally evaluate the performance of an existing argument mining system developed for essay scoring (named wLDA) when applied off-the-shelf to predict argument component labels for transcribed classroom discussions.
We then analyze the performance obtained when using the same features as wLDA to train a classifier specifically on our dataset.
We combine the wLDA feature set with features used in argument mining in the context of online dialogues and show that they are able to capture some of the similarities between online dialogues and our domain, and considerably improve the model.
We then evaluate two neural network models in several different scenarios pertaining to their input modality, the inclusion of handcrafted features, and the effect of multi-task learning when including specificity information.


\section{Related Work}
\label{sec:related_work}
With respect to the educational domain, previous studies in argument mining were largely aimed at student essays.
Persing and Ng \shortcite{Persing:15} studied argument strength with the ultimate goal of automated essay scoring.
Stab and Gurevych \shortcite{Stab:14} performed argument mining on student essays by first jointly performing argument component identification and classification, then predicting argument component relations.
Nguyen and Litman \shortcite{Nguyen:15} developed an argument mining system for analyzing student persuasive essays based on argument words and domain words. While domain words are used only in a specific topic, argument words are used across multiple topics and represent indicators of argumentative content. 
They later proposed an improved version of the system \shortcite{Nguyen:16}, which we will refer to as wLDA, by exploiting features able to abstract over specific essay topics and improve cross-topic performance.
While our current work is also aimed at developing argument mining systems in the educational context, we focus on educational discussion instead of student essays.
Our work also differs in the argument component types used:  we analyze claims, evidence, and warrants, while prior studies mostly focused on claims and premises.
The inclusion of warrants is particularly important to explicitly understand how students use them to connect evidence to claims.
As such, we do not expect prior models to work well on our corpus, although some of the features might still be useful.
Also, while some of the previously proposed systems address multiple subproblems simultaneously, e.g. argument component identification and argument component classification, we only focus on argument component classification.

Swanson et al. \shortcite{Swanson:15} developed a model for extracting argumentative portions of text from online dialogues, which were later used for summarizing the multiple argument facets.
Misra et al. \shortcite{Misra:15} analyzed dyadic online forum discussions to detect central propositions and argument facets.
Habernal and Gurevych \shortcite{Habernal:17} analyzed user-generated web discourse data from
several
sources 
by performing micro-level argumentation mining.
While these prior works analyze multi-party discussions, the discussions are neither  originally spoken nor in an educational context.

Like other areas of natural language processing, argument mining is experiencing an increase in the development of neural network models.
Niculae et al. \shortcite{Niculae:17} used a factor graph model which was parametrized by a recurrent neural network.
Daxenberger et al. \cite{Daxenberger:17} investigated the different conceptualizations of claims in several domains by analyzing in-domain and cross-domain performance of recurrent neural networks and convolutional neural networks, in addition to other models.
Schulz et al. \cite{Schulz:18} analyzed the impact of using multi-task learning when training  on a limited amount of labeled data.
In a similar way, we develop several convolutional neural network and recurrent neural network models, and also experiment with multi-task learning.
More detailed comparisons will be given in Section \ref{sec:argument_component_classification}.

\section{Dataset}
\label{sec:dataset}
We collected 73 transcripts of text-based classroom discussions, i.e. discussions centered on a text or literature piece (e.g. play, speech, book), for ELA high school level classes.
Some of the transcripts were gathered from published articles and dissertations, 
while the rest originated from videos which were transcribed by one of our annotators (see below).
While detailed demographic information for students participating in each discussion was not available, our dataset consists of a mix of small group (16 out of 73) versus whole class (57/73) discussions, both teacher-mediated (64/73) versus student only (9/73). Additionally, the discussions originated in urban schools (28/73), suburban schools (42/73), and schools located in small towns (3/73).

The unit of analysis for our work is argument move, which consists of a segment of text containing an argumentative discourse unit (ADU) \cite{Peldszus:13}.
Starting with transcripts broken down into turns at talk, an expert annotator segmented turns at talk into multiple argument moves when necessary: turns at talk containing multiple ADUs have been segmented into several argument moves, each consisting of a single ADU.
Turn segmentation effectively corresponds to argument component identification, and it is carried out manually.
We conducted a reliability study on turn segmentation with two annotators on a subset of the dataset consisting of 53 transcripts.
The reliability analysis resulted in Krippendorff $\alpha_U = 0.952$ \cite{Krippendorff:04}, which shows that turns at talk can be reliably segmented.

After segmentation, the data was manually annotated to capture two aspects of classroom talk, argument component and specificity, using the ELA classroom-oriented annotation scheme developed by Lugini et al. \shortcite{Lugini:18}.
The argument component types in this scheme, which is based on the Toulmin model \shortcite{Toulmin:58}, are:
$(i)$ {\it Claim}: an arguable statement that presents a particular interpretation of a text or topic.
$(ii)$ {\it Evidence\footnote{The ``evidence'' label is equivalent to ``data'' or ``grounds'' used in the original Toulmin model, though we use the label ``evidence'' to remain consistent with the annotation scheme.}}: facts, documentation, text reference, or testimony used to support or justify a claim.
$(iii)$ {\it Warrant}: reasons explaining how a specific evidence instance supports a specific claim.

Chisholm and Godley \shortcite{Chisholm:11} observed how specificity has an impact on the quality of the discussion, while Swanson et al. \shortcite{Swanson:15} noted that a relationship exists between specificity and the quality of arguments in online forum dialogues. For the purpose of investigating whether there exists a relationship between specificity and argument components, we additionally annotated data for specificity following the same coding scheme \cite{Lugini:18}.
Specificity labels are directly related to four elements for an argument move:
(1) it is specific to one (or a few) character or scene;
(2) it makes significant qualifications or elaborations;
(3) it uses content-specific vocabulary (e.g. quotes from the text);
(4) it provides a chain of reasons.
The specificity annotation scheme by Lugini et al. includes three labels along a linear scale:
$(i)$ {\it Low}: statement that does not contain any of these elements.
$(ii)$ {\it Medium}: statement that accomplishes one of these elements.
$(iii)$ {\it High}: statement that clearly accomplishes at least two specificity elements.
Only student turns were considered for annotations; teacher turns at talk were filtered out and do not appear in the final dataset.
Table \ref{tab:examples} shows a coded excerpt of a transcript from a discussion about the movie \textit{Princess Bride}.
\begin{table*}[t]
\centering
\begin{tabular}{|c|p{11cm}|c|c|}
\hline
\multicolumn{1}{|c|}{\textbf{Stu}} & \multicolumn{1}{c|}{\textbf{Argument Move}}                                                                                                                                                                                                                                                         & \multicolumn{1}{c|}{\textbf{Arg Comp}} & \multicolumn{1}{c|}{\textbf{Spec}} \\ \hline
S1                                & Well Fezzik went back to how he was,                                                                                                                                                                                                                                                                & Claim                                  & Low                                \\ \hline
S1                                & like how he gets lost. Then he goes like he needs to be around other people. And then finally when he does, he gets himself like relying on himself. But then right at the end, he doesn’t know where he’s at; he makes a wrong turn.                                                               & Evidence                                   & Med                                \\ \hline
S1                                & cause he tried doing it by himself and he can’t. So I think Fezzik went back to his normal ways, like after he changed.                                                                                                                                                                            & Warrant                                    & High                               \\ \hline
\end{tabular}
\caption{Coded excerpt of a discussion of the movie {\it Princess Bride}. Student S1 first makes a claim about Fezzik's behavior, then provides evidence by listing a series of events, then connects such events to his claim using a warrant. As the argument progresses, the specificity level increases.}
\label{tab:examples}
\vspace{-3mm}
\end{table*}

The resulting dataset consists of 2047 argument moves from 73 discussions.
As we can see from the label distribution shown in Table \ref{tab:data_statistics}, students 
produced a high number of claims, while warrant is the minority class.
\begin{table}[t]
\centering
\begin{tabular}{|c|c|c|}
\hline
\multicolumn{3}{|c|}{\textbf{Argument Component}} \\ \cline{1-3} 
Claim          & Warrant          & Evidence          \\ \hline
1034           & 358              & 655               \\ \cline{1-3} \hline \hline
\multicolumn{3}{|c|}{\textbf{Specificity}}             \\ \cline{1-3} 
Low            & Med              & High              \\ \cline{1-3} 
710            & 996              & 341               \\ \hline
\end{tabular}
\caption{Distribution of class labels for argument component type and specificity in our dataset. }
\label{tab:data_statistics}
\vspace{-3mm}
\end{table}
We can also observe a class imbalance for specificity labels, though the ratio between majority and minority classes is lower than that for argument component labels.

We evaluated inter-rater reliability on a subset of our dataset composed of 1049 argument moves from 50 discussions double-coded by two annotators.
Cohen's unweighted kappa for argument component labels was 0.629, while quadratic-weighted kappa for specificity labels (since they are ordered) was 0.641, which shows substantial agreement.


The average number of argument moves among the discussions is 27.3 while the standard deviation is 25.6, which shows a high variability in discussion length.
The average number of words per argument move and standard deviation are 22.6 and 22.1, respectively, which also shows large variability in how much students speak.

\section{Argument Component Classification}
\label{sec:argument_component_classification}
In this section we outline an existing argument component classification system that will serve as a baseline for our experiments, then propose several new models that use features extracted from neural networks and hand-crafted features, as well as models that use multi-task learning.

\subsection{Existing Argument Mining System}
\label{sec:wlda}
The wLDA\footnote{The original name of wLDA+4 stands for ``with LDA supported features and expanded with 4 features sets'' compared to their previous system. We use wLDA for brevity.} 
system was developed for performing argument component identification, classification, and relation extraction from student essays.
For the purpose of this study, we only consider the argument component classification subsystem.
The model is based on a support vector machine classifier which exploits features able to improve cross-topic performance.
The feature set consists of four main subsets: lexical features (argument words, verbs, adverbs, presence of modal verbs, discourse connectives, singular first person pronoun); parse features (argumentative subject-verb pairs, tense of the main verb, number of sub-clauses, depth of parse tree); structural features (number of tokens, token ratio, number of punctuation signs, sentence position, first/last paragraph, first/last sentence of paragraph); context features (number of tokens, number of punctuation signs, number of sub-clauses, modal verb in preceding/following sentences) extracted from the sentences before and after the one considered; four additional features for abstracting over essay topics.

Since the model was trained on essays annotated for major claim, claim, and premise, but not on warrants, in our evaluation we did not take into account misclassification errors for argument moves in our dataset labeled as warrants.
The pre-trained system performs argument component identification using a multiclass classification approach, such that each input will be classified as non argumentative, major claim, claim or premise.
Since our goal is to evaluate performance related to the component classification problem, we ignored all the argument moves classified as non argumentative by wLDA.
Considering the definitions of premise and evidence in the Toulmin model \shortcite{Toulmin:58}, we made the assumption of the two labels being equivalent for this study, i.e. if the predicted class for an argument move is premise and its gold standard label in our dataset is evidence, we consider the prediction correct.
In the same way we consider both claim and major claim labels as equivalent to claims in our dataset.

\subsection{Neural Network Models}
\begin{figure*}[t]
\setlength{\abovecaptionskip}{0pt plus 0pt minus 2pt}
\setlength{\belowcaptionskip}{-5pt plus 0pt minus 2pt}
\begin{center}
  \includegraphics[scale=0.41]{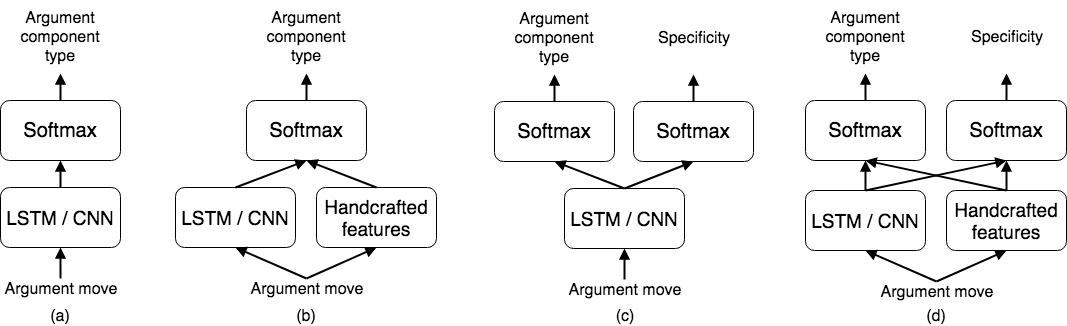}
  \end{center}
  \caption{Neural network models used in this study: neural network only setup (a); model incorporating neural network and handcrafted features (wLDA and online dialogue sets) (b); multi-task setup for neural network only model (c); multi-task setup for model using neural network and handcrafted features (d).
}
\label{fig:neural_network_models}
\vspace{-2mm}
\end{figure*}

Since the pre-trained model did not work well on our dataset, and the features it is based on show a large gap in performance compared to the original work (see Section \ref{sec:experiments}), we decided to use neural networks, and evaluate their ability to automatically extract meaningful features.
The proposed models consist of variations of two basic neural network models, namely Convolutional Neural Network (CNN) and Recurrent Neural Network (RNN) models. All the choices regarding the models were made in order to keep complexity and the number of weights at a minimum, since neural network models require in general a large amount of training data, while we have a limited size dataset.
The CNN model is based on a model proposed by Kim \shortcite{Kim:14cnn} and already used for argument mining in the past \cite{Daxenberger:17}, with a difference in the number of convolutional/pooling layers.
In particular, our model uses 3 convolutional/max pooling layers instead of 6, and only one fully connected layer after the convolutional ones, followed by a softmax layer used for classification.
This choice resulted from observing significant overfitting when increasing the number of convolutional layers due to the increase in the number of model weights  and the limited dataset size.
Figure \ref{fig:neural_network_models} shows diagrams for the different neural network setups used in our experiments.

The RNN model consists of a single Long-Short Term Memory (LSTM) network \cite{Hochreiter:97}. After propagating a complete argument move through the LSTM network, the resulting hidden state is the feature vector used as input to a softmax layer which outputs the predicted label.
Recurrent neural networks have also been used in the context of argument mining \cite{Daxenberger:17,Niculae:17}.
We set the size of the hidden state to 75 based on several factors. Following 
Bengio \cite{Bengio:12}, we decided to have an overcomplete network, i.e. one in which the size of the hidden state is bigger than the size of the input. Since the dimensionality of our character-based encoding is 37 and that for word-based embeddings is 50, we chose a hidden state with size greater than 50 (we use the same hidden state size for both models).
Increasing the  size introduced overfitting even quicker than the CNN model, given that the number of weights  increases more quickly for our LSTM model.

When using text as input to a neural network, we can generally view an argument move as either a sequence of characters, or as a sequence of words.
Unlike previous neural network-based argument mining models, each of our models was evaluated under both conditions: for character-based models we used a one-hot encoding (one-out of n) for each letter and number - special characters were filtered since they don't hold particular meaning in speech, and we cannot be sure of transcription conventions; for word-based models we used Global Vectors (GloVe) \cite{Pennington:14} with dimensionality of 50.
An important aspect to consider is that, while word-based models have some prior knowledge encoded in the word embeddings, character-based models do not.

Since neural network models usually require a large amount of training data to be effective, and we have relatively fewer number of argument moves compared to number of model weights, we also tested hybrid models in which a neural network output is combined with handcrafted features before the final softmax classification layer, as shown in Figure \ref{fig:neural_network_models} (b) and Figure \ref{fig:neural_network_models} (d).
Both CNN and LSTM models used categorical cross-entropy as loss function, and the number of epochs was automatically selected at training time by monitoring performance on a validation set consisting of 10\% of the training set for each fold.

\subsection{Multi-task Learning}
\begin{figure}[t]
\begin{center}
  \includegraphics[scale=0.47]{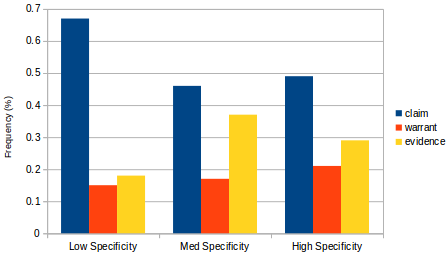}
  \end{center}
  \caption{Argument labels by specificity levels.}
  \label{fig:joint_label_distribution}
\vspace{-4mm}
\end{figure}

As we can see from Figure \ref{fig:joint_label_distribution}, the argument label distributions are different for the three specificity levels.
This leads us to believe a relationship exists between the specificity and argumentation annotations, therefore we decided to see whether specificity labels can be used to improve the performance of our argument mining models.

Multi-task learning for neural network models has shown promising results in the machine learning field \cite{Weston:12,Andrychowicz:16}.
It has recently been used in argument mining: Schulz et al. \shortcite{Schulz:18} proposed a multi-task learning setup in which the primary task consists of jointly performing argument component identification and classification (framed as a sequence tagging problem), while the additional tasks consist of the same task applied to different datasets. They showed that the multi-task models achieved better performance than single-task learning especially when limited in-domain training data is available for the primary task.

Unlike \cite{Schulz:18}, we decided to implement as secondary task specificity prediction on the same data as the primary task.
The underlying neural network setup was also different: while Schulz et al. used a bidirectional LSTM followed by a Conditional Random Field (CRF) classifier \cite{Reimers:17}, we were restricted to non-sequence classifiers.
We implemented multi-task learning in one of the standard ways: the embeddings generated by the networks are completely shared for both tasks of predicting argumentation and specificity.
For the CNN model, we added a second softmax layer for predicting specificity after the convolutional/pooling layers.
Similarly, for the LSTM model we added a second softmax layer that operates on the final hidden state of the network to predict specificity.
In both multi-task models specificity and argumentation are predicted at the same time, the loss function is computed as the sum of the individual loss functions for both tasks (the loss function for the specificity softmax layer is categorical cross-entropy as well), and gradient updates are backpropagated through the network.
This process results in embeddings trained jointly for the two tasks, which can effectively capture information relevant to both specificity and argumentation.

\subsection{Online Dialogue Features}
Since our dataset is based on multi-party discussion, it shares similarities with prior  argumentation work in multi-party online dialogues. Therefore we experiment with features from \cite{Swanson:15}, organized
into three main subsets: semantic-density features (number of pronouns, descriptive word-level statistics, number of occurrences of words of different lengths), lexical features (tf-idf feature for each unigram and bigram, descriptive argument move-level statistics), and syntactic features (unigrams, bigrams and trigrams of part of speech tags).
The only difference between the original features and the ones we implemented consists in the use of Speciteller \cite{Li:15}.
As observed by Lugini and Litman \cite{Lugini:17}, applying Speciteller as-is to domains other than news articles results in a considerable drop in performance.
Therefore, instead of including the specificity score obtained by directly applying Specificity to an argument move, we decided to use Speciteller's features.

\section{Experiments and Results}
\label{sec:experiments}
This section provides our experimental results.
In Section \ref{subsection:wlda_results} we will test our first hypothesis: using an argument mining system trained in a different domain will result in low performance, which can be improved by re-training on classroom discussions and by adding new features.
Section \ref{subsection:neural_nets_results} will be used to test our second hypothesis: neural network models can automatically extract important features for argument component classification.
Our third hypothesis will be tested in Section \ref{subsection:additional_features_results}: adding handcrafted features (i.e. online dialogue features, wLDA features) to the ones automatically extracted by neural networks will result in an increase of performance.
Lastly, we will test our fourth hypothesis in Section \ref{subsection:multitask_results}: jointly learning to predict argument component type and specificity will result in more robust models and achieve a further performance improvement.

Our experiments evaluate every model using a leave-one-transcript-out cross validation: each fold contains one transcript as test set and the remaining 72 as training set.
Cohen kappa, and unweighted precision, recall, and f-score were used as evaluation metrics.

The following python libraries were used for implementing and testing the different models: Scikit-learn \cite{scikit-learn},
Tensorflow \cite{tensorflow},
Keras \cite{keras},
NLTK \cite{nltk}.

Given that in our dataset warrants appear much less frequently than claims and evidence, data imbalance is a problem we need to address.
If trained naively, the limited amount of training data and the unbalanced class distribution lead the neural network models to specialize towards claims and evidence, with much weaker performance on warrants.
This is also the case for non neural network models, although the impact on performance is lower.
To combat this phenomenon we decided to use oversampling \cite{buda2017} in order to create a balanced dataset, hoping to further reduce the performance gap between the different classes
\footnote{We also tried setting class weights at training  to influence the loss function, though it only improved results marginally.}.
After computing the class frequency distribution on the training set, we randomly sampled moves from the two minority classes and added them to the current training set, repeating the process until  the class distribution was completely balanced (i.e. until the number of argument moves for each class equals the number of moves in the majority class)
\footnote{In the multi-task models oversampling was carried out only with respect to argument component labels since that is the primary task.},
while the test set was unchanged.

Table \ref{tab:results} shows the results for all  experiments.
The statistical significance  results in the table use the system in row 3 as the comparison baseline, as wLDA represents a system specifically designed for argument component classification (among other tasks). Additional statistical comparisons are provided in the text as well.

\subsection{Using wLDA Off the Shelf}
\label{subsection:wlda_results}
\begin{table*}[htbp]
\centering
\begin{tabular}{|c|p{3.5cm}|c|c|c|c||c|c|c|}
\hline
\textbf{Row} & \textbf{Models / Features}                                                                      & \textbf{Kappa} & \textbf{Precision} & \textbf{Recall} & \textbf{F-score} & \boldmath{$F_e$} & \boldmath{$F_w$} & \boldmath{$F_c$}\\ \hline
1 & Majority baseline                                                                                & 0.068          & 0.265              & 0.406           & 0.314      & 0.109 & 0.004 & 0.532      \\ \hline
2 & Pre-trained wLDA                                                                                & 0.077          & 0.289              & 0.350           & 0.269      & 0.351 & N/A & 0.456      \\ \hline
3 & Logistic Regression (wLDA features)                                                             & 0.142          & 0.412              & 0.394           & 0.379      & 0.390 & 0.211 & 0.540      \\ \hline
4 & Logistic Regression (wLDA features + online dialogue) & \textbf{0.283}          & 0.508              & 0.500           & 0.480      & 0.479 & 0.222 & \textbf{0.693}      \\ \hline
\hline
\multicolumn{9}{|l|}{\textit{Character level NN models}}                                                                                                                            \\ \hline
5 & LSTM                                                                                            & -0.002         & 0.062              & 0.253           & 0.082      & 0.007 & 0.242 & 0.013      \\ \hline
6 & LSTM + wLDA + online dialogue                                                                   & 0.034          & 0.217              & 0.304           & 0.150      & 0.080 & 0.272$^\ddagger$ & 0.090      \\ \hline
7 & CNN                                                                                             & 0.143          & 0.439              & 0.423           & 0.393     & 0.372 & 0.218 & 0.574       \\ \hline
8 & CNN + wLDA + online dialogue                                                                    & 0.241$^\star$          & 0.482              & 0.475           & 0.450      & 0.449 & 0.236 & 0.637      \\ \hline
\hline
\multicolumn{9}{|l|}{\textit{Word level NN models}}                                                                                                                                 \\ \hline
9 & LSTM                                                                                            & 0.069          & 0.408              & 0.399           & 0.218      & 0.161 & 0.198 & 0.295      \\ \hline
10 & LSTM + wLDA + online dialogue                                                                  & 0.181          & 0.462              & 0.447           & 0.391      & 0.362 & 0.279$^\ddagger$ & 0.522      \\ \hline
11 & CNN                                                                                             & 0.125          & 0.410              & 0.404           & 0.378      & 0.370 & 0.231 & 0.526      \\ \hline
12 & CNN + wLDA + online dialogue                                                                   & $0.241^\star$          & $0.492^\star$              & 0.488           & $0.455^\dagger$      & 0.468 & 0.276$^\ddagger$ & 0.622      \\ \hline
\hline
\multicolumn{9}{|l|}{\textit{Multi-task character level NN models}}                                                                                                                 \\ \hline
13 & LSTM                                                                                            & 0.060          & 0.408              & 0.399           & 0.208       & 0.134 & 0.203 & 0.287     \\ \hline
14 & LSTM + wLDA + online dialogue                                                                   & 0.117          & 0.379              & 0.375           & 0.287       & 0.362 & 0.279$^\ddagger$ & 0.522     \\ \hline
15 & CNN                                                                                             & 0.166          & 0.444              & 0.437           & 0.407       & 0.399 & 0.220 & 0.586     \\ \hline
16 & CNN + wLDA + online dialogue                                                                    & 0.259$^\dagger$          & 0.506$^\dagger$              & 0.488           & 0.468$^\star$       & 0.474 & 0.262$^\dagger$ & 0.640     \\ \hline
\hline
\multicolumn{9}{|l|}{\textit{Multi-task word level NN models}}                                                                                                                      \\ \hline
17 & LSTM                                                                                            & 0.093          & 0.379              & 0.364           & 0.276      & 0.298 & 0.252 & 0.378      \\ \hline
18 & LSTM + wLDA + online dialogue                                                                   & 0.232          & 0.497$^\dagger$              & 0.482           & 0.440     & 0.419 & 0.299$^\ddagger$ & 0.583       \\ \hline
19 & CNN                                                                                             & 0.164          & 0.351              & 0.443           & 0.441     & 0.476 & 0.249 & 0.598       \\ \hline
20 & CNN + wLDA + online dialogue                                                                    & $0.276^\ddagger$          & \textbf{0.521}$^\ddagger$              & \textbf{0.512}$^\dagger$           & \textbf{0.485}$^\dagger$       & \textbf{0.484} & \textbf{0.312}$^\ddagger$ & 0.638     \\ \hline
\end{tabular}
\caption{Results obtained with the baseline model/features and the proposed neural network models using different feature sets. Each line represents the average of a transcript-wise cross validation. Best results are in bold. $^\star$, $^\dagger$, and $^\ddagger$ indicate statistical significance at the 0.1, 0.05, and 0.01 levels respectively, compared to the model in row 3. The three right-most columns represent per-class F-score for evidence, warrants, and claims respectively.}
\label{tab:results}
\vspace{-4mm}
\end{table*}

Since not all the argument moves were considered when computing results for the pre-trained out of the box wLDA model (see Section \ref{sec:wlda}), the results in row 2 are not directly comparable to others. Nonetheless they show the upper bound in performance of the pre-trained model, and we can see that it is
comparable to
a majority baseline which always predicts the majority class in each fold.
This result shows that claims and evidence expressed in written essays and classroom discussions have very little in common.
This is clearer when we look at  improvement obtained training a logistic regression model\footnote{We also experimented with random forest, naive Bayes and support vector machines, but they provided inferior results compared to logistic regression.} using the same wLDA features on our dataset (row 3), which outperforms the pre-trained wLDA in all metrics (row 2), and indicates that the wLDA features are still able to somewhat distinguish between claims and evidence while performing considerably worse on warrants.
Additionally, if we add to this model the online dialogue feature set, the resulting model improves all results and obtains the best kappa overall (row 4).
This confirms our hypothesis: given the similarity that exists between our domain and online dialogues, features developed for analyzing argumentation in online dialogues are also useful in classroom discussions.

\subsection{Neural Network Models Alone}
\label{subsection:neural_nets_results}
Our second hypothesis is validated by the results in Table \ref{tab:results} by comparing row 3 with rows 7, 11, 15, and 19, where we can see that the CNN models achieve performance comparable to a classifier trained on features specifically developed for argument component classification. This indicates that convolutional neural network models are able to extract useful features.
Additionally, when comparing the best of these models (row 19, with respect to  f-score)  to the best performing model based only on handcrafted features (row 4), the difference in performance is not statistically significant for any of the metrics in Table \ref{tab:results}.

Looking more closely at the results obtained using neural network models alone we can see two different trends.
While LSTM models show performance comparable to random chance (e.g. row 5, with kappa close to zero and lower than the majority baseline), three of our  four CNN models (rows 7, 15, 19) perform as well as or better than the wLDA based model (row 3) (except for precision in row 19 and F$_e$ in row 7).
Overall, under the same conditions CNN models almost always outperform LSTM models.
One interesting difference between the two models is that the prior knowledge introduced by word embeddings in word-based models is essential for improving performance of LSTMs (e.g. row 5 vs row 9), while this is not the case for CNN models (e.g. row 7 vs row 11).
The length of sequences (i.e. argument moves) for character-based models makes it extremely hard for LSTMs to capture long-term dependencies, especially with limited amount of training data.
Convolutional models, on the other hand, learn kernels that effectively function as feature detectors and seem to be able to better distinguish important features, and do not always benefit from word level inputs.

\subsection{Adding wLDA Features and Online Dialogue Features}
\label{subsection:additional_features_results}
It is clear from Table \ref{tab:results} that almost all neural network models benefit from additional handcrafted features (with the exception of precision and recall for rows 13 and 14).
This is not surprising, given that neural networks require a large amount of data to be trained effectively, and although random oversampling helped, we still have a limited amount of training data.
Even when including additional features the two architectures show slightly different trends: CNN usually outperform LSTM, however LSTM models benefit more from the additional features. This is at least in part due to LSTMs initially having lower performance
without handcrafted features.
We analyzed the importance of different subsets of the online dialogue features through a feature ablation study.
For CNN models, removing any subset of features resulted in a decrease in performance, except for the \textit{syntax} subset in the \textit{character level CNN + wLDA + online dialogue} model in both single task and multi-task settings.
For LSTM models, all feature subsets contributed to increasing performance in the multi-task settings, while that was not always true for the single task models.

\subsection{Multi-task Learning}
\label{subsection:multitask_results}
Finally, we analyze the impact of multi-task learning in argument component classification.
Our findings are in line with the literature in other domains, with results showing that models trained on argumentation and specificity labels almost always outperform the ones trained only on argumentation.
LSTMs benefit from the multi-task setup more than CNN models: among all combinations of LSTM models, the only one able to achieve kappa greater than 0.2 and f-score greater than 0.4 is a multi-task one.
Additionally, the word-level CNN model using wLDA and online dialogue feature sets and trained using multi-task learning is the only model able to achieve f-score greater than 0.3 for warrants.

It should be noted that although the neural network based model at row 20 outperforms the logistic regression model at row 4 in terms of precision, recall, and F-score, the difference in performance is not statistically significant, and neither is the reduction in kappa and F$_c$.

\section{Conclusions and Future Work}
\label{sec:conclusions}
In this work we evaluated the performance of an existing argument mining system developed for a different educational application (i.e. student essays) on a corpus composed of spoken classroom discussions.
Although the pre-trained system showed poor performance on our dataset, its features show promising results when used in a model specifically trained on classroom discussions.
We extracted additional feature sets based on related work in the online dialogue domain, and showed that combining online dialogue and student essay features achieves the highest kappa on our dataset.
We then developed additional models based on two types of neural networks, showing that performance can be further improved.
We provided an experimental evaluation of the differences between convolutional networks and recurrent networks, and between character-based and word-based models.
Lastly, we showed that argument component classification models can benefit from multi-task learning, when adding a secondary task consisting of predicting specificity.

Even though we were able to achieve better performance compared to a pre-trained system and a majority baseline, we are far from the performance of argument mining systems in other domains such as student essays or legal texts.
Although the wLDA features extract information from previous argument moves, we plan to take advantage of the collaborative nature of our corpus by extending the feature sets in order to exploit contextual information and develop models that can explicitly take advantage of previous argument moves.
Given the performance improvements obtained with multi-task models, we also plan to extend these models and include additional tasks at training time with the hope of further boosting performance.
We also plan to add other types of cross validation, since leave-one-transcript-out introduces great variability in the composition of test sets, possibly attenuating the statistical significance for some results.

\section*{Acknowledgements}
We want to thank Amanda Godley, Christopher Olshefski, Tazin Afrin, Huy Nguyen, and Annika Swallen for their contribution, and all the anonymous reviewers for their helpful feedback.

This work was supported by the Learning Research and Development Center.

\bibliography{emnlp2018}

\begin{thebibliography}{43}
\expandafter\ifx\csname natexlab\endcsname\relax\def\natexlab#1{#1}\fi

\bibitem[{Abadi et~al.(2015)Abadi, Agarwal, Barham, Brevdo, Chen, Citro,
  Corrado, Davis, Dean, Devin, Ghemawat, Goodfellow, Harp, Irving, Isard, Jia,
  Jozefowicz, Kaiser, Kudlur, Levenberg, Man\'{e}, Monga, Moore, Murray, Olah,
  Schuster, Shlens, Steiner, Sutskever, Talwar, Tucker, Vanhoucke, Vasudevan,
  Vi\'{e}gas, Vinyals, Warden, Wattenberg, Wicke, Yu, and Zheng}]{tensorflow}
Mart\'{\i}n Abadi, Ashish Agarwal, Paul Barham, Eugene Brevdo, Zhifeng Chen,
  Craig Citro, Greg~S. Corrado, Andy Davis, Jeffrey Dean, Matthieu Devin,
  Sanjay Ghemawat, Ian Goodfellow, Andrew Harp, Geoffrey Irving, Michael Isard,
  Yangqing Jia, Rafal Jozefowicz, Lukasz Kaiser, Manjunath Kudlur, Josh
  Levenberg, Dandelion Man\'{e}, Rajat Monga, Sherry Moore, Derek Murray, Chris
  Olah, Mike Schuster, Jonathon Shlens, Benoit Steiner, Ilya Sutskever, Kunal
  Talwar, Paul Tucker, Vincent Vanhoucke, Vijay Vasudevan, Fernanda Vi\'{e}gas,
  Oriol Vinyals, Pete Warden, Martin Wattenberg, Martin Wicke, Yuan Yu, and
  Xiaoqiang Zheng. 2015.
\newblock {TensorFlow}: Large-scale machine learning on heterogeneous systems.
\newblock Software available from tensorflow.org.

\bibitem[{Andrychowicz et~al.(2016)Andrychowicz, Denil, Gomez, Hoffman, Pfau,
  Schaul, Shillingford, and De~Freitas}]{Andrychowicz:16}
Marcin Andrychowicz, Misha Denil, Sergio Gomez, Matthew~W Hoffman, David Pfau,
  Tom Schaul, Brendan Shillingford, and Nando De~Freitas. 2016.
\newblock Learning to learn by gradient descent by gradient descent.
\newblock In \emph{Advances in Neural Information Processing Systems}, pages
  3981--3989.

\bibitem[{Applebee et~al.(2003)Applebee, Langer, Nystrand, and
  Gamoran}]{Applebee:03}
Arthur~N Applebee, Judith~A Langer, Martin Nystrand, and Adam Gamoran. 2003.
\newblock Discussion-based approaches to developing understanding: Classroom
  instruction and student performance in middle and high school english.
\newblock \emph{American Educational Research Journal}, 40(3):685--730.

\bibitem[{Ashley and Walker(2013)}]{Ashley:13}
Kevin~D Ashley and Vern~R Walker. 2013.
\newblock Toward constructing evidence-based legal arguments using legal
  decision documents and machine learning.
\newblock In \emph{Proceedings of the Fourteenth International Conference on
  Artificial Intelligence and Law}, pages 176--180. ACM.

\bibitem[{Bengio(2012)}]{Bengio:12}
Yoshua Bengio. 2012.
\newblock Practical recommendations for gradient-based training of deep
  architectures.
\newblock In \emph{Neural networks: Tricks of the trade}, pages 437--478.
  Springer.

\bibitem[{Biber(1988)}]{Biber:88}
Douglas Biber. 1988.
\newblock \emph{Variation across speech and writing}.
\newblock Cambridge University Press.

\bibitem[{Bird et~al.(2009)Bird, Klein, and Loper}]{nltk}
Steven Bird, Ewan Klein, and Edward Loper. 2009.
\newblock \emph{Natural language processing with Python: analyzing text with
  the natural language toolkit}.
\newblock " O'Reilly Media, Inc.".

\bibitem[{Buda et~al.(2017)Buda, Maki, and Mazurowski}]{buda2017}
Mateusz Buda, Atsuto Maki, and Maciej~A Mazurowski. 2017.
\newblock A systematic study of the class imbalance problem in convolutional
  neural networks.
\newblock \emph{arXiv preprint arXiv:1710.05381}.

\bibitem[{Chafe and Tannen(1987)}]{Chafe:87}
W.~Chafe and D.~Tannen. 1987.
\newblock The relation between written and spoken language.
\newblock \emph{Annual Review of Anthropology}, 16(1):383--407.

\bibitem[{Chisholm and Godley(2011)}]{Chisholm:11}
James~S Chisholm and Amanda~J Godley. 2011.
\newblock Learning about language through inquiry-based discussion: Three
  bidialectal high school students’ talk about dialect variation, identity,
  and power.
\newblock \emph{Journal of Literacy Research}, 43(4):430--468.

\bibitem[{Chollet et~al.(2015)}]{keras}
Fran\c{c}ois Chollet et~al. 2015.
\newblock Keras.
\newblock \url{https://keras.io}.

\bibitem[{Daxenberger et~al.(2017)Daxenberger, Eger, Habernal, Stab, and
  Gurevych}]{Daxenberger:17}
Johannes Daxenberger, Steffen Eger, Ivan Habernal, Christian Stab, and Iryna
  Gurevych. 2017.
\newblock What is the essence of a claim? cross-domain claim identification.
\newblock In \emph{Proceedings of the 2017 Conference on Empirical Methods in
  Natural Language Processing}, pages 2055--2066. Association for Computational
  Linguistics.

\bibitem[{Elizabeth et~al.(2012)Elizabeth, Ross~Anderson, Snow, and
  Selman}]{Elizabeth:12}
Tracy Elizabeth, Trisha~L Ross~Anderson, Elana~H Snow, and Robert~L Selman.
  2012.
\newblock Academic discussions: An analysis of instructional discourse and an
  argument for an integrative assessment framework.
\newblock \emph{American Educational Research Journal}, 49(6):1214--1250.

\bibitem[{Engle and Conant(2002)}]{Engle:02}
Randi~A Engle and Faith~R Conant. 2002.
\newblock Guiding principles for fostering productive disciplinary engagement:
  Explaining an emergent argument in a community of learners classroom.
\newblock \emph{Cognition and Instruction}, 20(4):399--483.

\bibitem[{Ghosh et~al.(2014)Ghosh, Muresan, Wacholder, Aakhus, and
  Mitsui}]{Ghosh:14}
Debanjan Ghosh, Smaranda Muresan, Nina Wacholder, Mark Aakhus, and Matthew
  Mitsui. 2014.
\newblock Analyzing argumentative discourse units in online interactions.
\newblock In \emph{Proceedings of the First Workshop on Argumentation Mining},
  pages 39--48.

\bibitem[{Habernal and Gurevych(2017)}]{Habernal:17}
Ivan Habernal and Iryna Gurevych. 2017.
\newblock Argumentation mining in user-generated web discourse.
\newblock \emph{Computational Linguistics}, 43(1):125--179.

\bibitem[{Hochreiter and Schmidhuber(1997)}]{Hochreiter:97}
Sepp Hochreiter and J{\"u}rgen Schmidhuber. 1997.
\newblock Long short-term memory.
\newblock \emph{Neural computation}, 9(8):1735--1780.

\bibitem[{Kim(2014)}]{Kim:14cnn}
Yoon Kim. 2014.
\newblock Convolutional neural networks for sentence classification.
\newblock In \emph{Proceedings of the 2014 Conference on Empirical Methods in
  Natural Language Processing (EMNLP)}, pages 1746--1751.

\bibitem[{Krippendorff(2004)}]{Krippendorff:04}
Klaus Krippendorff. 2004.
\newblock Measuring the reliability of qualitative text analysis data.
\newblock \emph{Quality and Quantity}, 38:787--800.

\bibitem[{Lawrence and Reed(2017)}]{Lawrence:17}
John Lawrence and Chris Reed. 2017.
\newblock Using complex argumentative interactions to reconstruct the
  argumentative structure of large-scale debates.
\newblock In \emph{Proceedings of the 4th Workshop on Argument Mining}, pages
  108--117.

\bibitem[{Li and Nenkova(2015)}]{Li:15}
Junyi~Jessy Li and Ani Nenkova. 2015.
\newblock Fast and accurate prediction of sentence specificity.
\newblock In \emph{Proceedings of the Twenty-Ninth Conference on Artificial
  Intelligence (AAAI)}, pages 2281--2287.

\bibitem[{Lugini and Litman(2017)}]{Lugini:17}
Luca Lugini and Diane Litman. 2017.
\newblock Predicting specificity in classroom discussion.
\newblock In \emph{Proceedings of the 12th Workshop on Innovative Use of NLP
  for Building Educational Applications}, pages 52--61.

\bibitem[{Lugini et~al.(2018)Lugini, Litman, Amanda, and
  Christopher}]{Lugini:18}
Luca Lugini, Diane Litman, Godley Amanda, and Olshefski Christopher. 2018.
\newblock Annotating stdent talk in text-based classroom discussions.
\newblock In \emph{Proceedings of the 13th Workshop on Innovative Use of NLP
  for Building Educational Applications}, pages 110--116.

\bibitem[{McLaren et~al.(2010)McLaren, Scheuer, and
  Mik{\v{s}}{\'a}tko}]{Mclaren:10}
Bruce~M McLaren, Oliver Scheuer, and Jan Mik{\v{s}}{\'a}tko. 2010.
\newblock Supporting collaborative learning and e-discussions using artificial
  intelligence techniques.
\newblock \emph{International Journal of Artificial Intelligence in Education},
  20(1):1--46.

\bibitem[{Misra et~al.(2015)Misra, Anand, Fox~Tree, and Walker}]{Misra:15}
Amita Misra, Pranav Anand, Jean Fox~Tree, and Marilyn Walker. 2015.
\newblock Using summarization to discover argument facets in dialog.
\newblock In \emph{Proceedings of the 2015 Conference of the North American
  Chapter of the Association for Computational Linguistics: Human Language
  Technologies}.

\bibitem[{Murphy et~al.(2009)Murphy, Wilkinson, Soter, Hennessey, and
  Alexander}]{Murphy:09}
P~Karen Murphy, Ian~AG Wilkinson, Anna~O Soter, Maeghan~N Hennessey, and John~F
  Alexander. 2009.
\newblock Examining the effects of classroom discussion on students’
  comprehension of text: A meta-analysis.
\newblock \emph{Journal of Educational Psychology}, 101(3):740.

\bibitem[{Nguyen and Litman(2015)}]{Nguyen:15}
Huy Nguyen and Diane Litman. 2015.
\newblock Extracting argument and domain words for identifying argument
  components in texts.
\newblock In \emph{Proceedings of the 2nd Workshop on Argumentation Mining},
  pages 22--28.

\bibitem[{Nguyen and Litman(2016)}]{Nguyen:16}
Huy Nguyen and Diane~J Litman. 2016.
\newblock Improving argument mining in student essays by learning and
  exploiting argument indicators versus essay topics.
\newblock In \emph{FLAIRS Conference}, pages 485--490.

\bibitem[{Nguyen and Litman(2018)}]{Nguyen:18}
Huy~V Nguyen and Diane~J Litman. 2018.
\newblock Argument mining for improving the automated scoring of persuasive
  essays.
\newblock In \emph{Proceedings of the Thirty-Second AAAI Conference on
  Artificial Intelligence}.

\bibitem[{Niculae et~al.(2017)Niculae, Park, and Cardie}]{Niculae:17}
Vlad Niculae, Joonsuk Park, and Claire Cardie. 2017.
\newblock {Argument Mining with Structured SVMs and RNNs}.
\newblock In \emph{Proceedings of ACL}.

\bibitem[{Palau and Moens(2009)}]{Palau:09}
Raquel~Mochales Palau and Marie-Francine Moens. 2009.
\newblock Argumentation mining: the detection, classification and structure of
  arguments in text.
\newblock In \emph{Proceedings of the 12th international conference on
  artificial intelligence and law}, pages 98--107. ACM.

\bibitem[{Pedregosa et~al.(2011)Pedregosa, Varoquaux, Gramfort, Michel,
  Thirion, Grisel, Blondel, Prettenhofer, Weiss, Dubourg, Vanderplas, Passos,
  Cournapeau, Brucher, Perrot, and Duchesnay}]{scikit-learn}
F.~Pedregosa, G.~Varoquaux, A.~Gramfort, V.~Michel, B.~Thirion, O.~Grisel,
  M.~Blondel, P.~Prettenhofer, R.~Weiss, V.~Dubourg, J.~Vanderplas, A.~Passos,
  D.~Cournapeau, M.~Brucher, M.~Perrot, and E.~Duchesnay. 2011.
\newblock Scikit-learn: Machine learning in {P}ython.
\newblock \emph{Journal of Machine Learning Research}, 12:2825--2830.

\bibitem[{Peldszus and Stede(2013)}]{Peldszus:13}
Andreas Peldszus and Manfred Stede. 2013.
\newblock From argument diagrams to argumentation mining in texts: A survey.
\newblock \emph{International Journal of Cognitive Informatics and Natural
  Intelligence (IJCINI)}, 7(1):1--31.

\bibitem[{Pennington et~al.(2014)Pennington, Socher, and
  Manning}]{Pennington:14}
Jeffrey Pennington, Richard Socher, and Christopher Manning. 2014.
\newblock Glove: Global vectors for word representation.
\newblock In \emph{Proceedings of the 2014 conference on empirical methods in
  natural language processing (EMNLP)}, pages 1532--1543.

\bibitem[{Persing and Ng(2015)}]{Persing:15}
Isaac Persing and Vincent Ng. 2015.
\newblock Modeling argument strength in student essays.
\newblock In \emph{Proceedings of the 53rd Annual Meeting of the Association
  for Computational Linguistics and the 7th International Joint Conference on
  Natural Language Processing (Volume 1: Long Papers)}, volume~1, pages
  543--552.

\bibitem[{Reimers and Gurevych(2017)}]{Reimers:17}
Nils Reimers and Iryna Gurevych. 2017.
\newblock Argumentation mining in user-generated web discourse.
\newblock In \emph{Proceedings of the 2017 Conference on Empirical Methods in
  Natural Language Processing}, pages 338--348.

\bibitem[{Reznitskaya and Gregory(2013)}]{Reznitskaya:13}
Alina Reznitskaya and Maughn Gregory. 2013.
\newblock Student thought and classroom language: Examining the mechanisms of
  change in dialogic teaching.
\newblock \emph{Educational Psychologist}, 48(2):114--133.

\bibitem[{Schulz et~al.(2018)Schulz, Eger, Daxenberger, Kahse, and
  Gurevych}]{Schulz:18}
Claudia Schulz, Steffen Eger, Johannes Daxenberger, Tobias Kahse, and Iryna
  Gurevych. 2018.
\newblock Multi-task learning for argumentation mining in low-resource
  settings.
\newblock In \emph{Proceedings of the 2018 Conference of the North American
  Chapter of the Association for Computational Linguistics: Human Language
  Technologies, Volume 2 (Short Papers)}, pages 35--41. Association for
  Computational Linguistics.

\bibitem[{Stab and Gurevych(2014)}]{Stab:14}
Christian Stab and Iryna Gurevych. 2014.
\newblock Annotating argument components and relations in persuasive essays.
\newblock In \emph{Proceedings of COLING 2014, the 25th International
  Conference on Computational Linguistics: Technical Papers}, pages 1501--1510.

\bibitem[{Stab and Gurevych(2017)}]{Stab:17}
Christian Stab and Iryna Gurevych. 2017.
\newblock Parsing argumentation structures in persuasive essays.
\newblock \emph{Computational Linguistics}, 43(3):619--659.

\bibitem[{Swanson et~al.(2015)Swanson, Ecker, and Walker}]{Swanson:15}
Reid Swanson, Brian Ecker, and Marilyn Walker. 2015.
\newblock Argument mining: Extracting arguments from online dialogue.
\newblock In \emph{Proceedings of the 16th Annual Meeting of the Special
  Interest Group on Discourse and Dialogue}, pages 217--226.

\bibitem[{Toulmin(1958)}]{Toulmin:58}
Stephen Toulmin. 1958.
\newblock \emph{The uses of argument}.
\newblock Cambridge: Cambridge University Press.

\bibitem[{Weston et~al.(2012)Weston, Ratle, Mobahi, and Collobert}]{Weston:12}
Jason Weston, Fr{\'e}d{\'e}ric Ratle, Hossein Mobahi, and Ronan Collobert.
  2012.
\newblock Deep learning via semi-supervised embedding.
\newblock In \emph{Neural Networks: Tricks of the Trade}, pages 639--655.
  Springer.

\end{thebibliography}
\bibliographystyle{acl_natbib_nourl}

\end{document}